\def\year{2019}\relax
\documentclass[letterpaper]{article} 
\usepackage{aaai19}  
\usepackage{times}  
\usepackage{helvet}  
\usepackage{courier}  
\usepackage{url}  
\usepackage{graphicx}  
\usepackage{amsmath}
\usepackage{booktabs}
\usepackage{latexsym}
\usepackage{bm}
\usepackage{color}

\begin{document}
%
\title{Story Ending Generation with Incremental Encoding and Commonsense Knowledge}

\author{Jian Guan$^2$\thanks{~~Authors contributed equally to this work.} , Yansen Wang$^{1*}$, Minlie Huang$^1$\thanks{~~Corresponding author: Minlie Huang.}\\
    $^1$Dept. of Computer Science \& Technology, Tsinghua University, Beijing 100084, China\\
    $^1$Institute for Artificial Intelligence£¬Tsinghua University (THUAI), China\\
    $^1$Beijing National Research Center for Information Science and Technology, China\\
    $^2$Dept. of Physics, Tsinghua University, Beijing 100084, China\\
  {\tt guanj15@mails.tsinghua.edu.cn;ys-wang15@mails.tsinghua.edu.cn;}\\
  {\tt aihuang@tsinghua.edu.cn}}

\date{}
\maketitle
\begin{abstract}
    Generating a reasonable ending for a given story context, i.e., story ending generation, is a strong indication of story comprehension. This task requires not only to understand the context clues which play an important role in planning the plot, but also to  handle implicit knowledge to make a reasonable, coherent story.
    In this paper, we devise a novel model for story ending generation. The model adopts an incremental encoding scheme to represent context clues which are spanning in the story context. In addition, commonsense knowledge is applied through multi-source attention to facilitate story comprehension, and thus to help generate coherent and reasonable endings. Through building context clues and using implicit knowledge, the model is able to produce reasonable story endings.
     Automatic and manual evaluation shows that our model can generate more reasonable story endings than state-of-the-art baselines. \footnote{Our codes and data are available at \url{https://github.com/JianGuanTHU/StoryEndGen}.}
\end{abstract}
\section{Introduction}
    Story generation is an important but challenging task because it requires to deal with logic and implicit knowledge 
    \cite{Li2013Story,Von2016Generate,Ji2017Dynamic,Jain2017Story,Lara2018Event,Elizabeth2018Neural}. Story ending generation aims at concluding a story and completing the plot given a story context. We argue that solving this task involves addressing the following issues: 1) Representing the \textbf{context clues} which contain key information for planning a reasonable ending;
    and 2) Using \textbf{implicit knowledge} (e.g., commonsense knowledge) to facilitate understanding of the story and better predict what will happen next.

    \begin{figure}[!ht]
    \centering
    \includegraphics[width=3.2in]{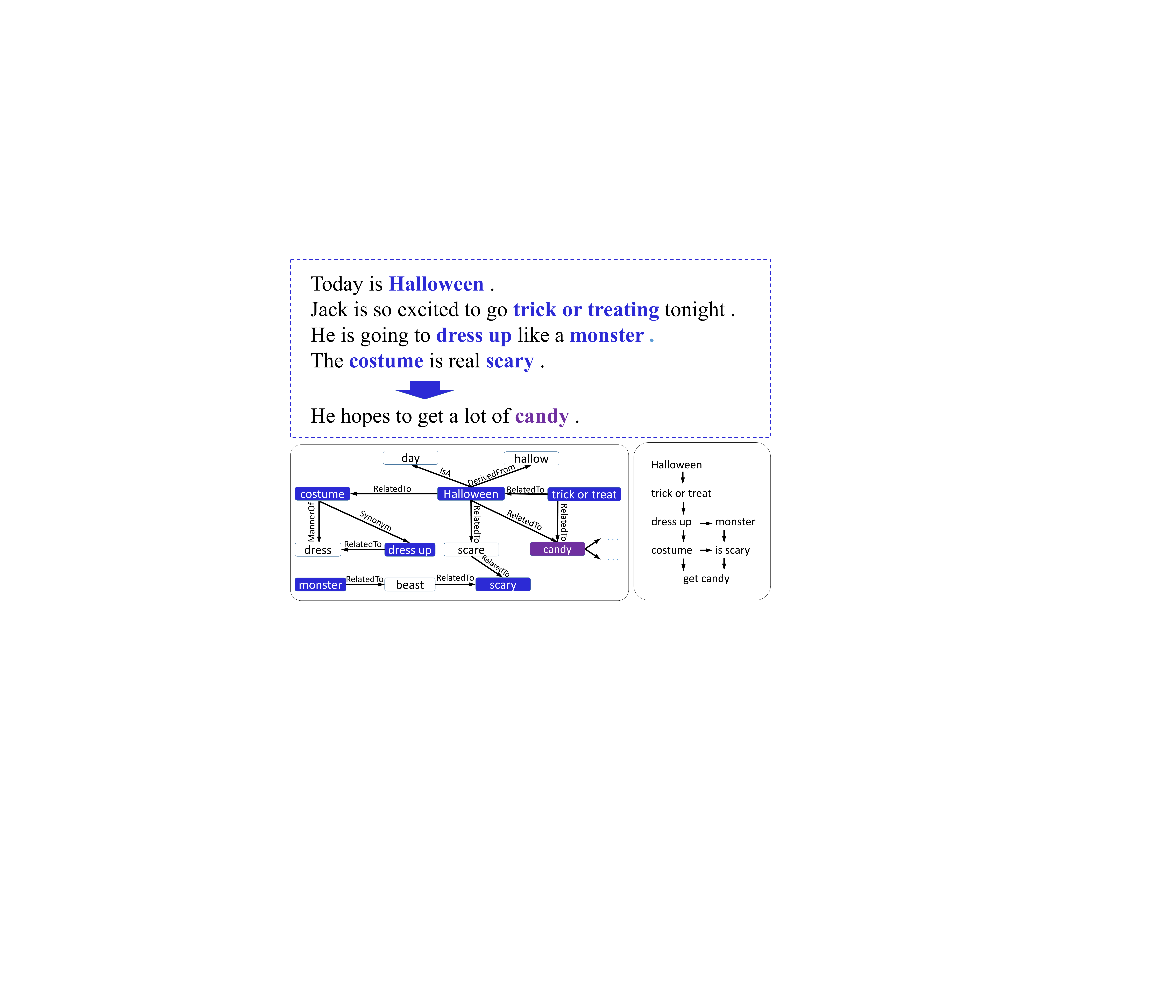}
    \caption{A story example. Words in blue/purple are events and entities. The bottom-left graph is retrieved from ConceptNet and the bottom-right graph represents how events and entities form the context clue.}
    \label{example}
    \end{figure}

   Comparing to textual entailment or reading comprehension~
   \cite{Dagan2006te,Hermann2015rc} 
   story ending generation requires more to deal with the logic and causality information that may span multiple sentences in a story context. The logic information in story can be captured by the appropriate sequence of events\footnote{Event in this paper refers to a verb or simple action such as {\it dress up, trick or treating} as shown in Figure \ref{example}.} or entities occurring in a sequence of sentences, and the chronological order or causal relationship between events or entities. 
    The ending should be generated from the whole context clue rather than merely inferred from a single entity or the last sentence. It is thus important for story ending generation to represent the context clues for predicting what will happen in an ending.

    However, deciding a reasonable ending not only depends on representing the context clues properly, but also on the ability of language understanding with implicit knowledge that is beyond the text surface. Humans use their own experiences and implicit knowledge to help understand a story. As shown in Figure \ref{example}, the ending talks about {\it candy} which can be viewed as commonsense knowledge about {\it Halloween}. Such knowledge can be crucial for story ending generation.

    Figure \ref{example} shows an example of a typical story in the ROCStories corpus~\cite{Mostafazadeh2016Story}. In this example, the events or entities in the story context constitute the context clues which reveal the logical or causal relationships between events or entities. These concepts, including {\it Halloween}, {\it trick or treat}, and {\it monster}, are connected as a graph structure. A reasonable ending should consider all the connected concepts rather than just some individual one. Furthermore, with the help of commonsense knowledge retrieved from ConceptNet~\cite{Speer2012Representing}, it is easier to infer a reasonable ending with the knowledge that {\it candy} is highly related to {\it Halloween}.

    To address the two issues in story ending generation, we devise a model that is equipped with an \textbf{incremental encoding scheme} to encode context clues effectively, 
    and a \textbf{multi-source attention mechanism} to use \textit{commonsense knowledge}.
    The representation of the context clues is built through incremental reading (or encoding) of the sentences in the story context one by one. When encoding a current sentence in a story context, the model can attend not only to the words in the preceding sentence, but also the knowledge graphs which are retrieved from ConceptNet for each word. In this manner, commonsense knowledge can be encoded in the model through graph representation techniques, and therefore, be used to facilitate understanding story context and inferring coherent endings.
    Integrating the context clues and commonsense knowledge, the model can generate more reasonable endings than state-of-the-art baselines.

    Our contributions are as follows:
    \begin{itemize}
        \item To assess the machinery ability of story comprehension, we investigate story ending generation from two new angles. One is to modeling logic information via incremental context clues and the other is the utility of implicit knowledge in this task.

        \item We propose a neural model which represents context clues by incremental encoding, and leverages commonsense knowledge by multi-source attention, to generate logical and reasonable story endings. Results show that the two techniques are effective in capturing the coherence and logic of story.

    \end{itemize}
\section{Related Work}
    The corpus we used in this paper was first designed for Story Cloze Test (SCT)~\cite{Mostafazadeh2016A}, which requires to select a correct ending from two candidates given a story context. Feature-based~
    \cite{Chaturvedi2017Story,Lin2017Reasoning}
    or neural \cite{Mostafazadeh2016Story,Wang2017Conditional} classification models are proposed to measure the coherence between a candidate ending and a story context from various aspects such as event, sentiment, and topic. However, story ending generation \cite{C18-1088,Zhao2018blcu,Peng2018Towards} is more challenging in that the task requires to modeling context clues and implicit knowledge to produce reasonable endings.

    Story generation, moving forward to complete story comprehension, is approached as selecting a sequence of events to form a story by satisfying a set of criteria~\cite{Li2013Story}. Previous studies can be roughly categorized into two lines: rule-based methods and neural models. Most of the traditional rule-based methods for story generation ~\cite{Li2013Story,Von2016Generate} retrieve events from a knowledge base with some pre-specified semantic relations.

    Neural models for story generation has been widely studied with sequence-to-sequence (seq2seq) learning~\cite{roemmele2016seq2seq}. And various contents such as photos and independent descriptions are largely used to inspire the story~
    \cite{Jain2017Story}.To capture the deep meaning of key entities and events, \citeauthor{Ji2017Dynamic}~(\citeyear{Ji2017Dynamic}) and \citeauthor{Elizabeth2018Neural}~(\citeyear{Elizabeth2018Neural}) explicitly modeled the entities mentioned in story with dynamic representation, and \citeauthor{Lara2018Event}~(\citeyear{Lara2018Event}) decomposed the problem into planning successive events and generating sentences from some given events. \citeauthor{angela2018HSG}~(\citeyear{angela2018HSG}) adopted a hierarchical architecture to generate the whole story from some given keywords.

     Commonsense knowledge is beneficial for many natural language tasks such as semantic reasoning and text entailment, which is particularly important for story generation. \citeauthor{Peter2011Types}~(\citeyear{Peter2011Types}) characterized the types of commonsense knowledge mostly involved in recognizing textual entailment. 
     Afterwards, commonsense knowledge was used in natural language inference ~\cite{Samuel2015Large,Sheng2017Ordinal} and language generation~\cite{Zhou2018Commonsense}. 
     \citeauthor{Todor2018kr}~(\citeyear{Todor2018kr}) incorporated external commonsense knowledge into a neural cloze-style reading comprehension model. \citeauthor{Hannah2018event2mind}~(\citeyear{Hannah2018event2mind}) performed commonsense inference on people's intents and reactions of the event's participants given a short text. Similarly, \citeauthor{Rashkin2018Modeling}~(\citeyear{Rashkin2018Modeling}) introduced a  new annotation framework to explain psychology of story characters with commonsense knowledge. And commonsense knowledge has also been shown useful to choose a correct story ending from two candidate endings ~\cite{Lin2017Reasoning,li2018mann}.

\section{Methodology}
\subsection{Overview}

    \begin{figure*}[!ht]
    \centering
    \includegraphics[width=6.3in]{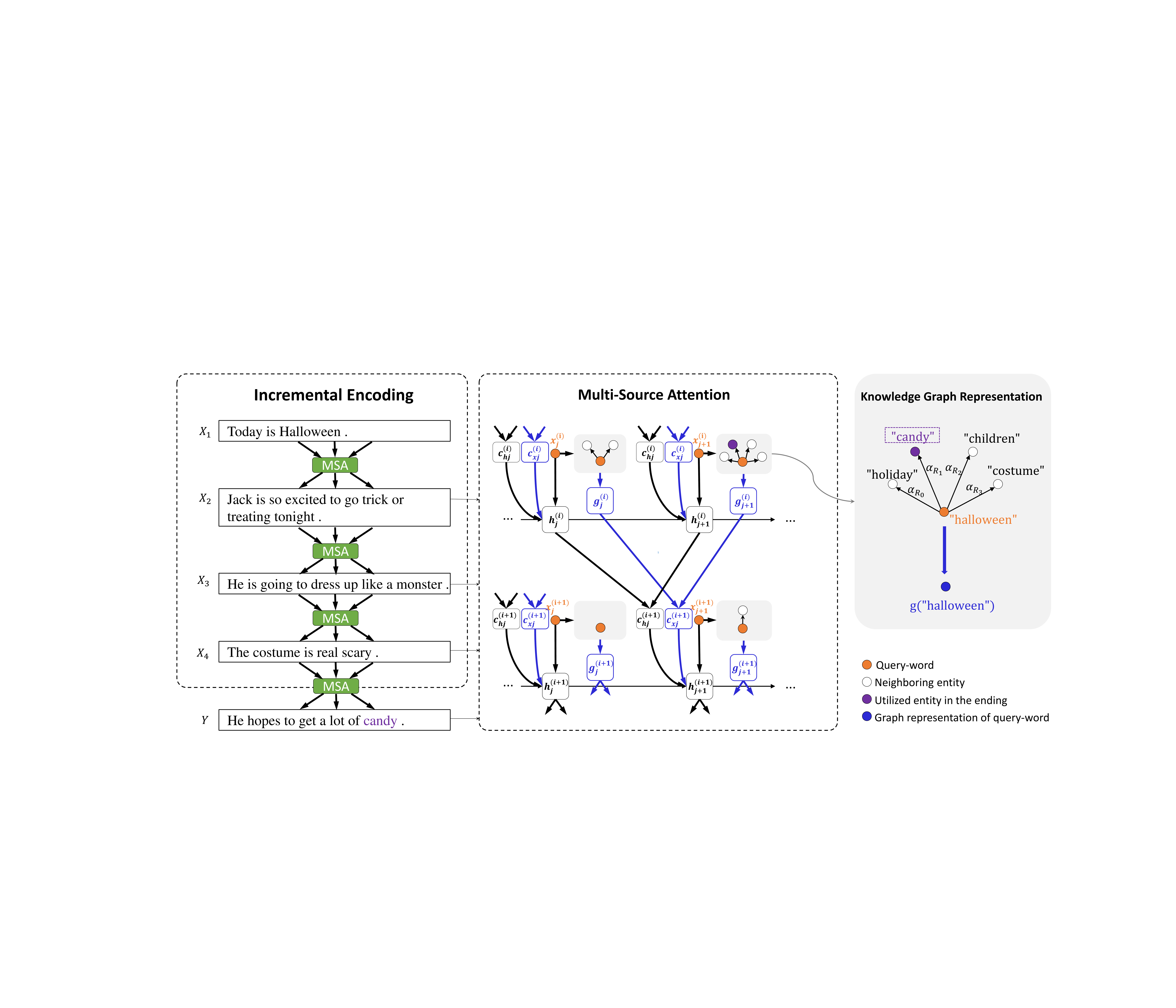}
    \caption{Model overview. The model is equipped with incremental encoding (IE) and multi-source attention (MSA). $x_j^{(i)}$: the $j$-th word in sentence $i$; $\mathbf{c}_{hj}^{(i)}$: state context vector; $\mathbf{c}_{xj}^{(i)}$: knowledge context vector; $\mathbf{g}_j^{(i)}$: graph vector of word $x_j^{(i)}$; $\mathbf{h}_{j}^{(i)}$: $j$-th hidden state of sentence $i$. The state (knowledge) context vectors are attentive read of hidden states (graph vectors) in the preceding sentence.
    }
    \label{model_ove}
    \end{figure*}

    The task of story ending generation can be stated as follows: given a story context consisting of a sentence sequence $X=\{X_1, X_2, \cdots, X_K\}$\footnote{Adjacent sentences in story context have much logic connection, temporal dependency, and casual relationship.}, where $X_i=x_1^{(i)}x_2^{(i)}\cdots x_{l_i}^{(i)}$ represents the $i$-th sentence containing $l_i$ words, the model should generate a one-sentence ending $Y=y_1y_2...y_l$ which is reasonable in logic, formally as
    \begin{align}
        {Y^*} = \mathop{argmax}\limits_{Y} \mathcal{P}(Y|X).
    \end{align}

    As aforementioned, context clue and commonsense knowledge is important for modeling the logic and casual information in story ending generation. To this end, we devise an \textbf{incremental encoding scheme} based on the general encoder-decoder framework \cite{sutskever2014sequence}. As shown in Figure \ref{model_ove}, the scheme encodes the sentences in a story context incrementally with a \textbf{multi-source attention mechanism}: when encoding sentence $X_{i}$, the encoder obtains a context vector which is an attentive read of the hidden states, and the graph vectors of the preceding sentence $X_{i-1}$. In this manner, the relationship between words (some are entities or events) in sentence $X_{i-1}$ and those in $X_{i}$ is built {\it incrementally}, and therefore, the chronological order or causal relationship between entities (or events) in adjacent sentences can be captured implicitly. To leverage commonsense knowledge which is important for generating a reasonable ending, a one-hop knowledge graph for each word in a sentence is retrieved from ConceptNet, and each graph can be represented by a vector in two ways. The incremental encoder not only attends to the hidden states of $X_{i-1}$, but also to the graph vectors at each position of $X_{i-1}$. By this means, our model can generate more reasonable endings by representing context clues and encoding commonsense knowledge.


\subsection{Background: Encoder-Decoder Framework}
    The encoder-decoder framework is a general framework widely used in text generation. Formally, the model encodes the input sequence $X=x_1x_2\cdots x_m$ into a sequence of hidden states, as follows,
    \begin{align}
        \textbf{h}_{t} &= \mathbf{LSTM}(\textbf{h}_{t-1}, \bm{e}(x_t)), \label{general-encoding}
    \end{align}
    where $\textbf{h}_{t}$ denotes the hidden state at step $t$ and $\bm{e}(x)$ is the word vector of $x$.

    At each decoding position, the framework will generate a word by sampling from the word distribution $\mathcal{P}(y_t|y_{<t},X)$ ($y_{<t}=y_1y_2\cdots y_{t-1}$ denotes the sequences that has been generated before step $t$), which is computed as follows:
    \begin{align}
            &\mathcal{P}(y_t|y_{<t}, X) = \mathbf{softmax}(\textbf{W}_{0}\bm{s}_{t}+\textbf{b}_0), \label{S2Sposibility}\\
            &\textbf{s}_{ t} = \mathbf{LSTM}(\textbf{s}_{ t-1}, \bm{e}(y_{t-1}), \textbf{c}_{t-1}), \label{S2Sdecoder}
    \end{align}
    where $\textbf{s}_t$ denotes the decoder state at step $t$. When an attention mechanism is applied, $\textbf{c}_{t-1}$ is an attentive read of the context, which is a weighted sum of the encoder's hidden states as $\textbf{c}_{t-1}=\sum_{i=1}^m\alpha_{(t-1)i}\textbf{h}_i$, and $\alpha_{(t-1)i}$ measures the association between the decoder state $\textbf{s}_{t-1}$ and the encoder state $\textbf{h}_i$. Refer to \cite{Bahdanau2014Neural} for more details.

\subsection{Incremental Encoding Scheme}
    Straightforward solutions for encoding the story context can be: 1) Concatenating the $K$ sentences to a long sentence and encoding it with an LSTM ; or 2) Using a hierarchical LSTM with hierarchical attention \cite{Yang2016Hierarchical}, which firstly attends to the hidden states of a sentence-level LSTM, and then to the states of a word-level LSTM. However, these solutions are not effective to represent the context clues which may capture the key logic information. Such information revealed by the chronological order or causal relationship between events or entities in adjacent sentences.

    To better represent the context clues, we propose an incremental encoding scheme: when encoding the current sentence $X_i$, it obtains a context vector which is an attentive read of the preceding sentence $X_{i-1}$. In this manner, the order/relationship between the words in adjacent sentences can be captured implicitly.

    This process can be stated formally as follows:
    \begin{align}
        \textbf{h}_{j}^{(i)} = \mathbf{LSTM}(\textbf{h}_{j-1}^{(i)}, \bm{e}(x_j^{(i)}), \textbf{c}_{\textbf{l}j}^{(i)}), ~i\ge 2. \label{incremental-encoding}
    \end{align}
    where $\textbf{h}^{(i)}_{j}$ denotes the hidden state at the $j$-th position of the $i$-th sentence, $\bm{e}(x_j^{(i)})$ denotes the word vector of the $j$-th word $x_j^{(i)}$. $\textbf{c}_{\textbf{l},j}^{(i)}$ is the context vector which is an attentive read of the \textit{preceding} sentence $X_{i-1}$, conditioned on $\textbf{h}^{(i)}_{j-1}$. We will describe the context vector in the next section.

    During the decoding process, the decoder obtains a context vector from the last sentence $X_{K}$ in the context to utilize the context clues. The hidden state is obtained as below:
    \begin{align}
        &\textbf{s}_{t} = \mathbf{LSTM}(\textbf{s}_{t-1}, \bm{e}(y_{t-1}), \textbf{c}_{\textbf{l}t}), \label{incremental-decoder}
    \end{align}
    where $\textbf{c}_{\textbf{l}t}$ is the context vector which is the attentive read of the last sentence $X_K$, conditioned on $\textbf{s}_{t-1}$. More details of the context vector will be presented in the next section.


\subsection{Multi-Source Attention (MSA)}
    The context vector ($\textbf{c}_{\textbf{l}}$) plays a key role in representing the context clues because it captures the relationship between words (or states) in the current sentence and those in the preceding sentence.
    As aforementioned, story comprehension sometime requires the access of implicit knowledge that is beyond the text.
    Therefore, the context vector consists of two parts, computed with multi-source attention. The first one $\textbf{c}_{\textbf{h}j}^{(i)}$ is derived by attending to the hidden states of the preceding sentence, and the second one $\textbf{c}_{\textbf{x}j}^{(i)}$ by attending to the knowledge graph vectors which represent the one-hop graphs in the preceding sentence. The MSA context vector is computed as follows:
    \begin{align}
        \textbf{c}_{\textbf{l}j}^{(i)} = \textbf{W}_\textbf{l}([\textbf{c}_{\textbf{h}j}^{(i)}; \textbf{c}_{\textbf{x}j}^{(i)}])+\textbf{b}_\textbf{l},
        \label{eq:context-vector}
    \end{align}
    where $\oplus$ indicates vector concatenation.
    Hereafter, $\textbf{c}_{\textbf{h}j}^{(i)}$ is called {\it state context vector}, and $\textbf{c}_{\textbf{x}j}^{(i)}$ is called {\it knowledge context vector}.

    The {\bf state context vector} is a weighted sum of the hidden states of the preceding sentence $X_{i-1}$ and can be computed as follows:
    \begin{align}
        \textbf{c}_{\textbf{h}j}^{(i)} &= \sum_{k = 1}^{l_{i-1}}\alpha_{h_k,j}^{(i)}\textbf{h}_{k}^{(i-1)}, \label{source-state-vector}\\
        \alpha_{h_k,j}^{(i)} &= \frac{e^{\beta_{h_k,j}^{(i)}}}{\;\sum\limits_{m=1}^{l_{i-1}}e^{\beta_{h_m,j}^{(i)}}\;},\\
        \beta_{h_k,j}^{(i)} &= \textbf{h}_{j-1}^{(i)\rm T}\textbf{W}_\textbf{s} \textbf{h}_k^{(i-1)},
    \end{align}
    where $\beta_{h_k,j}^{(i)}$ can be viewed as a weight between hidden state $\textbf{h}_{j-1}^{(i)}$ in sentence $X_i$  and hidden state $\textbf{h}_k^{(i-1)}$ in the preceding sentence $X_{i-1}$.


    Similarly, the {\bf knowledge context vector} is a weighted sum of the {\bf graph vectors} for the preceding sentence. Each word in a sentence will be used as a query to retrieve a one-hop commonsense knowledge graph from ConceptNet, and then, each graph will be represented by a {\bf graph vector}. After obtaining the graph vectors, the knowledge context vector can be computed by:
    \begin{align}
        \textbf{c}_{\textbf{x}j}^{(i)} &= \sum_{k = 1}^{l_{i-1}}\alpha_{x_k,j}^{(i)}\textbf{g}(x_{k}^{(i-1)}), \label{commonsense-vector}\\
        \alpha_{x_k,j}^{(i)} &= \frac{e^{\beta_{x_k,j}^{(i)}}}{\;\sum\limits_{m=1}^{l_{i-1}}e^{\beta_{x_m,j}^{(i)}}\;},\\
        \beta_{x_k,j}^{(i)} &= \textbf{h}_{j-1}^{(i)\rm T}\textbf{W} _\textbf{k}\textbf{g}(x_k^{(i-1)}),
    \end{align}
    where $\textbf{g}(x_k^{(i-1)})$ is the {\bf graph vector} for the graph which is retrieved for word $x_k^{(i-1)}$. Different from $\bm{e}(x_k^{(i-1)})$ which is the word vector, $\textbf{g}(x_k^{(i-1)})$ encodes commonsense knowledge and  extends the semantic representation of a word through neighboring entities and relations.

    During the decoding process, the knowledge context vectors are similarly computed by attending to the last input sentence $X_K$. There is no need to attend to all the context sentences because the context clues have been propagated within the incremental encoding scheme.


\subsection{Knowledge Graph Representation}
    Commonsense knowledge can facilitate language understanding and generation. To retrieve commonsense knowledge for story comprehension, we resort to ConceptNet\footnote{https://conceptnet.io}~\cite{Speer2012Representing}. ConceptNet is a semantic network which consists of triples $R=(h, r, t)$ meaning that head concept $h$ has the relation $r$ with tail concept $t$. Each word in a sentence is used as a query to retrieve a one-hop graph from ConceptNet. The knowledge graph for a word extends (encodes) its meaning by representing the graph from neighboring concepts and relations. 

    There have been a few approaches to represent commonsense knowledge.
    Since our focus in this paper is on using knowledge to benefit story ending generation, instead of devising new methods for representing knowledge, we adopt two existing methods: 1) \textbf{graph attention} \cite{veli?kovi?2018graph,Zhou2018Commonsense}, and 2) \textbf{contextual attention}~\cite{Todor2018kr}.
    We compared the two means of knowledge representation in the experiment.

\subsubsection{Graph Attention}
    Formally, the knowledge graph of word (or concept) $x$ is represented by a set of triples, $\mathbf{G}(x)=\{R_1, R_2, \cdots, R_{N_x}\}$ (where each triple $R_i$ has the same head concept $x$), and the graph vector $\bm{g}(x)$ for word $x$ can be computed via {\it graph attention}, as below:
    \begin{align}
        \textbf{g}(x) &= \sum_{i = 1}^{N_x}\alpha_{R_i}[\textbf{h}_i ; \textbf{t}_i],\\
        \alpha_{R_i} &= \frac{e^{\beta_{R_i}}}{\;\sum\limits_{j=1}^{N_x}e^{\beta_{R_j}}\;},\\
        \beta_{R_i} =
        (\textbf{W}_{\textbf{r}}&\textbf{r}_i)^{\rm T}\mathop{tanh}(\textbf{W}_{\textbf{h}}\textbf{h}_i+\textbf{W}_{\textbf{t}}\textbf{t}_i),
    \end{align}
    where $(h_i, r_i, t_i) = R_i \in \mathbf{G}(x)$ is the $i$-th triple in the graph. We use word vectors to represent concepts, i.e. $\textbf{h}_i = \bm{e}(h_i), \textbf{t}_i = \bm{e}(t_i)$, and learn trainable vector $\textbf{r}_i$ for relation $r_i$, which is randomly initialized.

    Intuitively, the above formulation assumes that the knowledge meaning of a word can be represented by its neighboring concepts (and corresponding relations) in the knowledge base.
    Note that entities in ConceptNet are common words (such as {\it tree, leaf, animal}), we thus use word vectors to represent h/r/t directly, instead of using geometric embedding methods (e.g., TransE) to learn entity and relation embeddings. In this way, there is no need to bridge the representation gap between geometric embeddings and text-contextual embeddings (i.e., word vectors).

\subsubsection{Contextual Attention}
    When using {\it contextual attention},  the graph vector $\textbf{g}(x)$ can be computed as follows:
    \begin{align}
    \textbf{g}(x)&=\sum_{i=1}^{N_x}\alpha_{R_i}\textbf{M}_{R_i},\\
    \textbf{M}_{R_i}&=BiGRU(\textbf{h}_i,\textbf{r}_i,\textbf{t}_i),\\
    \alpha_{R_i} &= \frac{e^{\beta_{R_i}}}{\;\sum\limits_{j=1}^{N_x}e^{\beta_{R_j}}\;},\\
    \beta_{R_i}&= \textbf{h}_{(x)}^{\rm T}\textbf{W}_\textbf{c}\textbf{M}_{R_i},
    \end{align}
    where $\textbf{M}_{R_i}$ is the final state of a \textit{BiGRU} connecting the elements of triple $R_i$, which can be seen as the knowledge memory of the $i$-th triple, while $\textbf{h}_{(x)}$ denotes the hidden state at the encoding position of word $x$.




\subsection{Loss Function}
    As aforementioned, the incremental encoding scheme is central for story ending generation. To better model the chronological order and causal relationship between adjacent sentences, we impose supervision on the encoding network. At each encoding step, we also generate a distribution over the vocabulary, very similar to the decoding process:
    \begin{align}
        \mathcal{P}(y_t|y_{<t}, X) =\mathbf{softmax}(\textbf{W}_{0}\textbf{h}_{j}^{(i)}+\textbf{b}_0),
    \end{align}
    Then, we calculate the negative data likelihood as loss function:
    \begin{align}
        \Phi &= \Phi_{en} + \Phi_{de}\\
         \Phi_{en} &= \sum_{i=2}^K\sum_{j=1}^{l_i} - \log \mathcal{P}(x_j^{(i)}=\widetilde{x}_j^{(i)}|x_{<j}^{(i)}, X_{<i}),\\
        \Phi_{de} &= \sum_t - \log \mathcal{P}(y_t=\tilde{y}_t|y_{<t}, X),
    \end{align}
    where $\widetilde{x}_j^{(i)}$ means the reference word used for encoding at the $j$-th position in sentence $i$, and $\tilde{y}_t$ represents the $j$-th word in the reference ending. Such an approach does not mean that at each step there is only one correct next sentence, exactly as many other generation tasks. Experiments show that it is better in logic than merely imposing supervision on the decoding network.

\section{Experiments}
\subsection{Dataset}
    We evaluated our model on the ROCStories corpus \cite{Mostafazadeh2016A}. The corpus contains 98,162 five-sentence stories for evaluating story understanding and script learning. The original task is designed to select a correct story ending from two candidates, while our task is to generate a reasonable ending given a four-sentence story context. We randomly selected 90,000 stories for training and the left 8,162 for evaluation. The average number of words in $X_1/X_2/X_3/X_4/Y$ is 8.9/9.9/10.1/10.0/10.5 respectively. The training data contains 43,095 unique words, and 11,192 words appear more than 10 times. For each word, we retrieved a set of triples from ConceptNet and stored those whose head entity and tail entity are noun or verb, meanwhile both occurring in SCT. Moreover, we retained at most 10 triples if there are too many. The average number of triples for each query word is 3.4.

\subsection{Baselines}
    We compared our models with the following state-of-the-art baselines:

    \noindent\textbf{Sequence to Sequence (Seq2Seq):} A simple encoder-decoder model which concatenates four sentences to a long sentence with an attention mechanism \cite{luong2015effective}.

    \noindent\textbf{Hierarchical LSTM (HLSTM):} The story context is represented by a hierarchical LSTM: a word-level LSTM for each sentence and a sentence-level LSTM connecting the four sentences \cite{Yang2016Hierarchical}. A hierarchical attention mechanism is applied, which attends to the states of the two LSTMs respectively.

    \noindent\textbf{HLSTM+Copy:} The copy mechanism \cite{Gu2016Incorporating} is applied to hierarchical states to copy the words in the story context for generation.

    \noindent\textbf{HLSTM+Graph Attention(GA):} We applied multi-source attention HLSTM where commonsense knowledge is encoded by graph attention.

    \noindent\textbf{HLSTM+Contextual Attention(CA):} Contextual attention is applied to represent commonsense knowledge.




\subsection{Experiment Settings}
    The parameters are set as follows: GloVe.6B~\cite{Pennington2014Glove} is used as word vectors, and the vocabulary size is set to 10,000 and the word vector dimension to 200. We applied 2-layer LSTM units with 512-dimension hidden states. These settings were applied to all the baselines.

    The parameters of the LSTMs (Eq. \ref{incremental-encoding} and \ref{incremental-decoder}) are shared by the encoder and the decoder.

\subsection{Automatic Evaluation}
    We conducted the automatic evaluation on the 8,162 stories (the entire test set). We generated endings from all the models for each story context.

\subsubsection{Evaluation Metrics}
    We adopted {\it perplexity}(PPL) and  {\it BLEU }\cite{Papineni2002IBM} to evaluate the generation performance. Smaller perplexity scores indicate better performance.  {\it BLEU } evaluates $n$-gram overlap between a generated ending and a reference ending. However, since there is only one reference ending for each story context, BLEU scores will become extremely low for larger $n$. We thus experimented with $n=1,2$. Note also that there may exist multiple reasonable endings for the same story context.

\subsubsection{Results}

    \begin{table}[!ht]
    \scriptsize
    \centering
    \begin{tabular}{l c c c c c}
    \toprule
    Model & PPL & BLEU-1 & BLEU-2 & Gram. & Logic.\\
    \midrule
    Seq2Seq & 18.97 & 0.1864 & 0.0410 & 1.74 & 0.70\\
    HLSTM & 17.26 & 0.2459 & 0.0771 & 1.57 & 0.84\\
    HLSTM+Copy & 19.93 & 0.2469 & 0.0783 & 1.66 & 0.90\\
    HLSTM+MSA(GA) & 15.75 & 0.2588 & 0.0809 & 1.70 & 1.06\\
    HLSTM+MSA(CA) & 12.53 & 0.2514 & 0.0825 & 1.72 & 1.02\\
    \hline
    IE (ours) & 11.04 & 0.2514 & 0.0813 & \textbf{1.84} & 1.10\\
    IE+MSA(GA) (ours) & 9.72& 0.2566 & 0.0854 & 1.68 & \textbf{1.26}\\
    IE+MSA(CA) (ours) & \textbf{8.79} & \textbf{0.2682} & \textbf{0.0936} & 1.66 & 1.24 \\
    \bottomrule
    \end{tabular}
    \caption{Automatic and manual evaluation results.}
    \label{auto-eva}
    \end{table}

    The results of the automatic evaluation are shown in Table \ref{auto-eva}\footnote{The BLEU-2 results in the AAAI version is corrected in this version. Conclusions are not affected.}, where IE means a simple incremental encoding framework that ablated the knowledge context vector from $\textbf{c}_{\textbf{l}}$ in Eq. (\ref{eq:context-vector}). Our models have lower perplexity and higher BLEU scores than the baselines. IE and IE+MSA have remarkably lower perplexity than other models. As for BLEU, IE+MSA(CA) obtained the highest BLEU-1 and BLEU-2 scores. This indicates that multi-source attention leads to generate story endings that have more overlaps with the reference endings.

\subsection{Manual Evaluation}
    Manual evaluations are indispensable to evaluate the coherence and logic of generated endings. For manual evaluation, we randomly sampled 200 stories from the test set and obtained 1,600 endings from the eight models. Then, we resorted to Amazon Mechanical Turk~(MTurk) for annotation. Each ending will be scored by three annotators and majority voting is used to select the final label.

\subsubsection{Evaluation Metrics}
    We defined two metrics - {\it grammar} and {\it logicality} for manual evaluation. Score 0/1/2 is applied to each metric during annotation.
    \paragraph{Grammar~(Gram.):} Whether an ending is natural and fluent. Score 2 is for endings without any grammar errors, 1 for endings with a few errors but still understandable and 0 for endings with severe errors and incomprehensible.

    \paragraph{Logicality~(Logic.):} Whether an ending is reasonable and coherent with the story context in logic. Score 2 is for reasonable endings that are coherent in logic, 1 for relevant endings but with some discrepancy between an ending and a given context, and 0 for totally incompatible endings.

    Note that the two metrics are scored independently. To produce high-quality annotation, we prepared guidelines and typical examples for each metric score.

\subsubsection{Results}


    The results of the manual evaluation are also shown in Table \ref{auto-eva}. Note that the difference between IE and IE+MSA exists in that IE does not attend to knowledge graph vectors in a preceding sentence, and thus it does use any commonsense knowledge.
    The incremental encoding scheme without MSA obtained the best grammar score and our full mode IE+MSA(GA) has the best logicality score. All the models have fairly good grammar scores (maximum is 2.0), while the logicality scores differ remarkably, much lower than the maximum score, indicating the challenges of this task.

    More specifically, \textbf{incremental encoding is effective} due to the facts: 1) IE is significantly better than Seq2Seq and HLSTM in grammar (Sign Test, 1.84 vs. $1.74/1.57$, p-value=$0.046/0.037$, respectively), and in logicality (1.10 vs. 0.70/0.84, p-value$<0.001/0.001$). 2) IE+MSA is significantly better than HLSTM+MSA in logicality (1.26 vs. 1.06,  p-value=$0.014$ for GA; 1.24 vs. 1.02, p-value=$0.022$ for CA). This indicates that incremental encoding is more powerful than traditional (Seq2Seq) and hierarchical (HLSTM) encoding/attention in utilizing context clues. Furthermore, \textbf{using commonsense knowledge leads to significant improvements in logicality}. The comparison in logicality between IE+MSA and IE (1.26/1.24 vs. 1.10, p-value=$0.028/0.042$ for GA/CA, respectively), HLSTM+MSA and HLSTM (1.06/1.02 vs. 0.84, p-value$<0.001/0.001$ for GA/CA, respectively), and  HLSTM+MSA and HLSTM+Copy (1.06/1.02 vs. 0.90, p-value=$0.044/0.048$, respectively) all approve this claim. In addition, similar results between GA and CA show that commonsense knowledge is useful but multi-source attention is not sensitive to  the knowledge representation scheme.

    More detailed results are listed in Table \ref{score-ratio}. Comparing to other models, IE+MSA has a much larger proportion of endings that are good both in grammar and logicality ($2$-$2$). The proportion of good logicality (score=2.0) from IE+MSA is much larger than that from IE (45.0\%+5.0\%/41.0\%+4.0\% vs. 36.0\%+2.0\% for GA/CA, respectively), and also remarkable larger than those from other baselines. Further, HLSTM equipped with MSA is better than those without MSA, indicating that commonsense knowledge is helpful. And the kappa measuring inter-rater agreement is 0.29 for three annotators, which implies a fair agreement.
    \begin{table}[!ht]
    \small
    \centering
    \begin{tabular}{l c c c c}
    \toprule
    Gram.-Logic. Score & 2-2 & 2-1 & 1-2 & 1-1 \\
    \midrule
    Seq2seq & 20.0\% &     22.0\% &      6.5\% &      1.5\% \\
    HLSTM &     21.0\% &     17.0\% &     10.0\% &      3.5\% \\
    HLSTM+Copy &     28.0\% &     19.0\% &      7.0\% &      5.5\% \\
    HLSTM+MSA(GA) &     33.5\% &     25.0\% &      5.0\% &      4.0\%\\
    HLSTM+MSA(CA) & 30.0\%& 26.0\%&2.0\%& 8.0\%\\
    \midrule
    IE (ours) &     36.0\% &     \textbf{34.0\%} &      2.0\% &      4.0\% \\
    IE+MSA(GA) (ours) &     \textbf{45.0\%} &     24.0\% &      5.0\% &      2.0\% \\
    IE+MSA(CA) (ours) & 41.0\%& 27.0\%& 4.0\% &2.0\%\\
    \bottomrule
    \end{tabular}
    \caption{Data distribution over Gram.-Logic. scores. $a$-$b$ denotes that the grammar score is $a$ and the logicality score is $b$. Each cell denotes the proportion of the endings with score $a$-$b$.}
    \label{score-ratio}
    \end{table}

\subsection{Examples and Attention Visualization}
    We presented an example of generated story endings in Table \ref{cases}. Our model generates more natural and reasonable endings than the baselines.

    \begin{table}[!ht]
    \small
    \centering
    \begin{tabular}{p{1.3cm} p{5.6cm}}
    \toprule

    \textbf{Context:} & Martha is \textbf{{cooking}} a special \textbf {meal} for her family.\\
     & She \textbf{{wants everything to be just right}} for when they eat.\\
     & Martha \textbf{perfects everything} and puts her \textbf{{dinner}} into the \textbf{oven}.\\
     & Martha goes to \textbf{lay down} for a quick \textbf{nap}.\\
    \textbf{Golden Ending:}& She \textbf{\underline {oversleeps}} and runs into the \textbf{\underline {kitchen}} to take out her \textbf{\underline {burnt dinner}}.\\
    \midrule
    \textbf{Seq2Seq:} & She was so happy to have a \textbf{\textit{new cake}}.\\
    \textbf{HLSTM:} & Her family \textbf{\textit{and her family}} are very happy with her \textbf{\underline {food}}.\\
    \textbf{HLSTM+ Copy:} & \textbf{\underline {Martha}} is happy to be able to \textbf{\textit{eat her family}}.\\
    \textbf{HLSTM+ GA:} & She is happy to be able to \textbf{\underline{cook her dinner}}.\\
    \textbf{HLSTM+ CA:}&She is very happy that she has made a new \textbf{\underline {cook}} .\\
    \midrule
    \textbf{IE:} & She is very happy with her \textbf{\underline {family}}.\\
    \textbf{IE+GA:} & When she gets back to the \textbf{\underline {kitchen}}, she sees a \textbf{\underline {burning light}} on the \textbf{\underline {stove}}.\\
    \textbf{IE+CA:} &She realizes the \textbf{\underline {food}} and is happy she was ready to \textbf{\underline {cook}} .\\
    \bottomrule
    \end{tabular}
    \caption{Generated endings from different models. \textbf{Bold} words denote the \textbf{key} entity and event in the story. \textit{Improper} words in ending is in \textit{italic} and \underline{proper} words are \underline{underlined}.}
    \label{cases}
    \end{table}
    In this example, the baselines predicted  wrong events in the ending. Baselines (Seq2Seq, HLSTM, and  HLSTM+Copy) have predicted improper entities (\textit{cake}), generated repetitive contents (\textit{her family}), or copied wrong words (\textit{eat}). The models equipped with incremental encoding or knowledge through MSA(GA/CA) perform better in this example. The ending by IE+MSA is more coherent in logic, and fluent in grammar. We can see that there may exist multiple reasonable endings for the same story context.

\begin{figure}[!ht]
        \centering
        \includegraphics[width=3.5in]{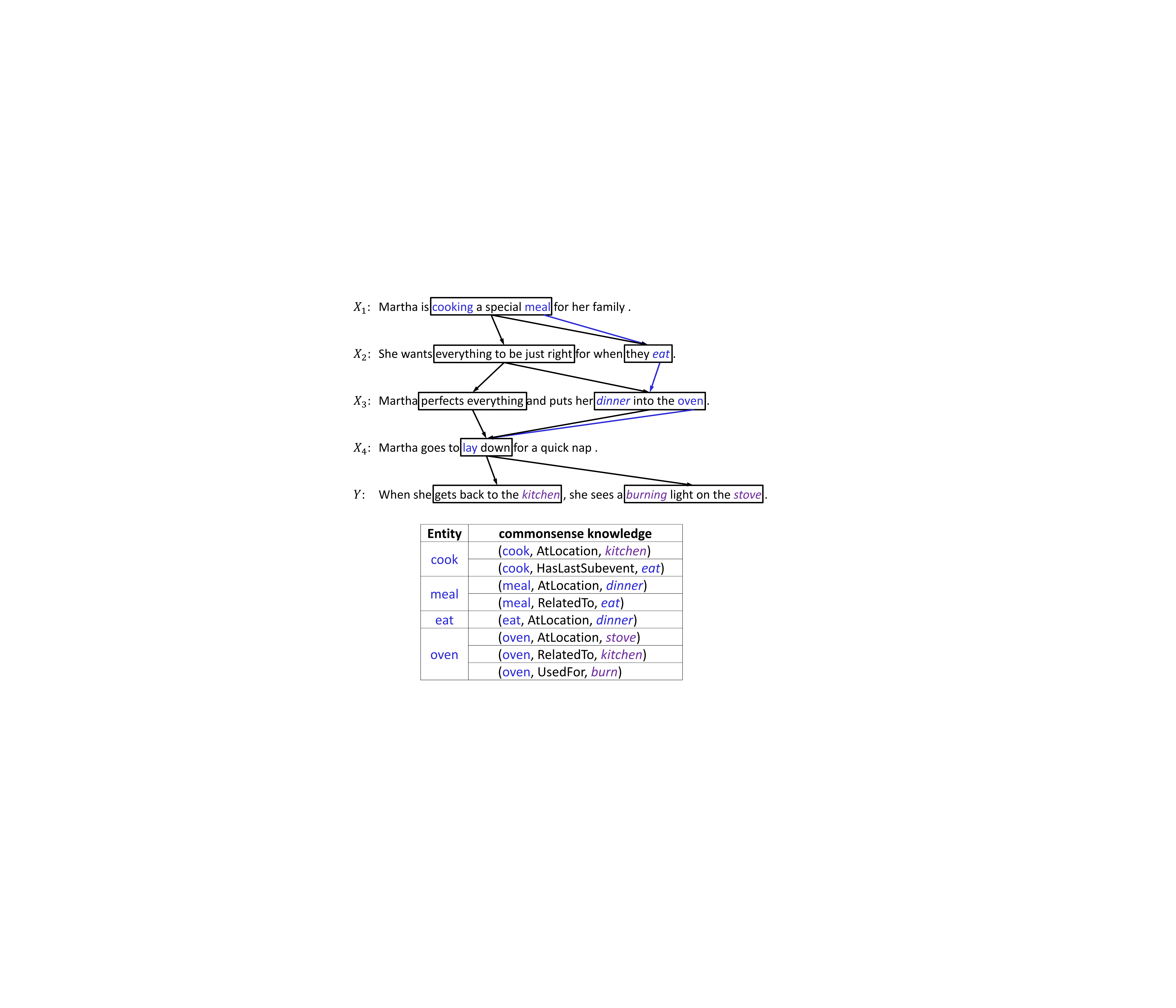}
        \caption{An example illustrating how incremental encoding builds connections between context clues. }
        \label{visual_att}
    \end{figure}

    In order to verify the ability of our model to utilize the context clues and implicit knowledge when planning the story plot, we visualized the attention weights of this example, as shown in Figure \ref{visual_att}. Note that this example is produced from graph attention.

    In Figure \ref{visual_att}, phrases in the box are key events of the sentences that are manually highlighted. Words in blue or purple are entities that can be retrieved from ConceptNet, respectively in story context or in ending.
    An arrow indicates that the words in the current box (e.g., {\it they eat} in $X_2$) all have largest attention weights to some words in the box of the preceding sentence (e.g., {\it cooking a special meal} in $X_1$). Black arrows are for state context vector (see Eq.\ref{source-state-vector}) and blue for knowledge context vector (see Eq.\ref{commonsense-vector}). For instance, {\it eat} has the largest knowledge attention to {\it meal} through the knowledge graph ($<${\it meal, AtLocation, dinner}$>$,$<${\it meal, RelatedTo, eat}$>$). Similarly, {\it eat} also has the second largest attention weight to {\it cooking} through the knowledge graph. For attention weights of state context vector, both words in {\it perfects everything} has the largest weight to some of {\it everything to be just right} ({\it everything}$\to${\it everything}, {\it perfect} $\to$ {\it right}).

    The example illustrates how the connection between context clues are built through incremental encoding and use of commonsense knowledge. The chain of context clues, such as ${be\_cooking}\to{want\_everything\_be\_right}\to{perfect\_everything}\to{lay\_down}\to{get\_back}$, and the commonsense knowledge, such as $<$\textit{cook, AtLocation, kitchen}$>$ and $<$\textit{oven, UsedFor, burn}$>$, are useful for generating reasonable story endings.

\section{Conclusion and Future Work}
    We present a story ending generation model that builds context clues via incremental encoding and leverages commonsense knowledge with multi-source attention. It encodes a story context incrementally with a multi-source attention mechanism to utilize not only context clues but also commonsense knowledge: when encoding a sentence, the model obtains a multi-source context vector which is an attentive read of the words and the corresponding knowledge graphs of the preceding sentence in the story context.
    Experiments show that our models can generate more coherent and reasonable story endings.

    As future work, our incremental encoding and multi-source attention for using commonsense knowledge may be applicable to other language generation tasks.

    \textbf{Refer to the Appendix for more details.}

\section{Acknowledgements}
    This work was jointly supported by the National Science Foundation of China  (Grant No.61876096/61332007), and the National Key R\&D Program of China (Grant No. 2018YFC0830200). We would like to thank Prof. Xiaoyan Zhu for her generous support.

\bibliographystyle{aaai}
\bibliography{emnlp2018}

\subsection{Appendix A: Annotation Statistics}
    We presented the statistics of annotation agreement in Table \ref{anno-stat}. The proportion of the annotations in which at least two annotators (3/3+2/3) assigned the same score to an ending is 96\% for grammar and 94\% for logicality. We can also see that the 3/3 agreement for logicality is much lower than that for grammar, indicating that logicality is more complicated for annotation than grammar.
    \begin{table}[!ht]
    \centering
    \begin{tabular}{l c c c}
    \toprule
    Metric & 3/3 & 2/3 & 1/3 \\
    \midrule
    Gram. & 58\% & 38\% & 4\%\\
    Logic. & 38\% & 57\% & 6\% \\
    \bottomrule
    \end{tabular}
    \caption{Annotation statistics. $n/3$ agreement means $n$ out of 3 annotators assigned the same score to an ending.}
    \label{anno-stat}
    \end{table}

\subsection{Appendix B: Error Analysis}
    We analyzed error types by manually checking all 46 bad endings generated by our model, where {\it bad} means the average score in terms of at least one metric is not greater than 1.

    There are 3 typical error types: {\it bad grammar} (BG), {\it bad logicality} (BL), and other errors. The distribution of types is shown in Table \ref{wrong-type}.

    \begin{table}[!ht]
    \centering
    \begin{tabular}{l c c c}
    \toprule
    \textbf{Error Type} & BG & BL & Others\\
    \midrule
    \textbf{IE+MGA} & 32.6\% & 37.0\% & 30.4\%  \\
    \bottomrule
    \end{tabular}
    \caption{Distribution of errors.}
    \label{wrong-type}
    \end{table}

    \begin{table}[!ht]
    \small
    \centering
    \begin{tabular}{p{1.2cm} p{5.6cm}}
    \toprule
    \textbf{Context:} & Sean \textbf{practiced driving} every day.\\
     & He \textbf{studied hard} for the written part of the \textbf{driving test}.\\
     & Sean felt ready to \textbf{take his driving test} after a couple months.\\
     & Sean was confident that he would \textbf{pass his driving test}.\\
     \textbf{Golden Ending:} & Sean \textbf{\underline{passed his driving test}} on the first attempt .\\
    \textbf{IE+MSA (GA):} & He \textbf{\underline{failed his test}} \textbf{\textit{and failed his test}}.\\
    \midrule
    \textbf{Context:} & Danny wanted to give his \textbf{daughter} a sweet \textbf{gift}.\\
     & DJ was \textbf{turning 18 years old} and ready to \textbf{drive}.\\
     & Danny decided it was time for \textbf{a new car}.\\
     & Danny went down to the dealership and \textbf{bought a new car}.\\
    \textbf{Golden Ending:} & He \textbf{\underline{surprised}} DJ on her \textbf{\underline{18th birthday}} with a \textbf{\underline{new car}}.\\
    \textbf{IE+MSA (GA):} & He \textbf{\underline{was so happy with}} his \textbf{\textit{new daughter}}.\\
    \midrule
    \textbf{Context:} & There was a big \textbf{pothole} near Bill's house.\\
     & He had to \textbf{swerve} to avoid it all the time.\\
     & He kept \textbf{calling the city} to \textbf{fix} it.\\
     & Eventually someone \textbf{returned his calls}.\\
     \textbf{Golden Ending:} & It took three more months for it to \textbf{\underline{be fixed}} \\
    \textbf{IE+MSA (GA):} & He was \textbf{\textit{relieved}} to find that he had to \textbf{\underline{work on his own}}.\\
    \bottomrule
    \end{tabular}
    \caption{Typical cases of errors. Bold words denote the key entity and event in the story. Improper words in ending is in italic and proper is underlined.}
    \label{error-case}
    \end{table}

    We also presented some typical cases for each error type in Table \ref{error-case}. Note that we only took \textit{graph attention} as example. The first case shows an instance of bad grammar for repetitive generation. The second case shows that our model predicted a wrong entity at the last position where {\it car} is obviously more appropriate than {\it daughter}. It happens when the attention focuses on the wrong position, but in more cases it happens due to the noise of the commonsense knowledge base\footnote{Note that ConceptNet contains many low-quality triples.}. The ending of the third case contains a relevant event {\it work on his own} but the event is not consistent to the previous word {\it relieved}.
    Other cases show that our model is not good at dealing with rare words. However, this can be further improved by applying copy mechanism, as our future work. These errors also indicate that story ending generation is challenging, and logic and implicit knowledge plays a central role in this task.

\subsection{Appendix C: Attention Visualization}
    The multi-source attention mechanism computes the state context vectors and knowledge context vectors respectively as follows:
    \begin{align}
        \textbf{c}_{\textbf{h}j}^{(i)} &= \sum_{k = 1}^{l_{i-1}}\alpha_{h_k,j}^{(i)}\textbf{h}_{k}^{(i-1)}, \label{source-state-vector}\\
        \textbf{c}_{\textbf{x}j}^{(i)} &= \sum_{k = 1}^{l_{i-1}}\alpha_{x_k,j}^{(i)}\textbf{g}(x_{k}^{(i-1)}), \label{commonsense-vector}
    \end{align}

    \begin{figure}[!ht]
        \centering
        \includegraphics[width=3.0in]{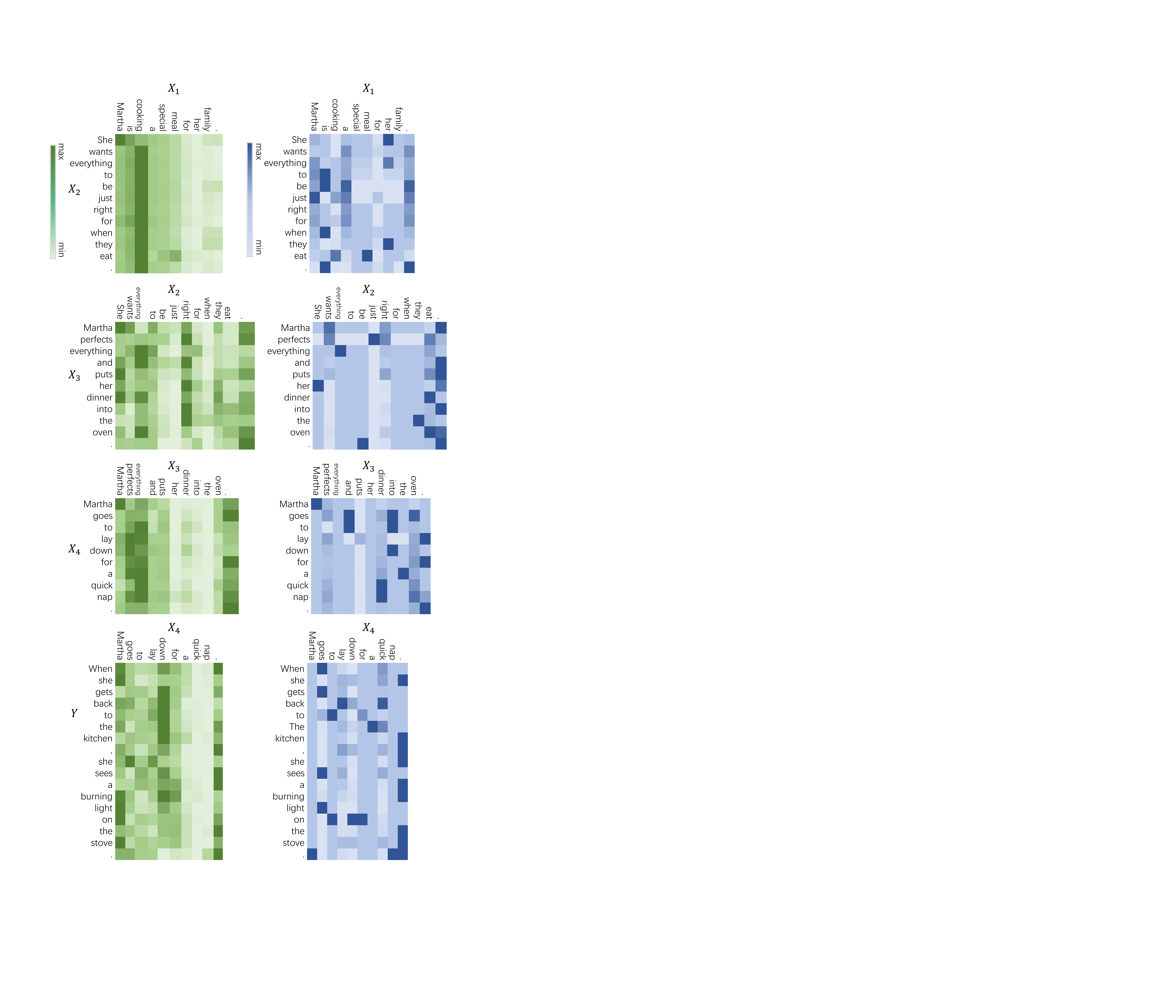}
        \caption{Attention weight visualization by multi-source attention. Green graphs are for state context vectors (Eq. \ref{source-state-vector}) and blue for knowledge context vectors (Eq. \ref{commonsense-vector}). Each graph represents the attention weight matrix for two adjacent sentences ($X_i$ in row and $X_{i+1}$ in column). The five graphs in the left column show the state attention weights in the incremental encoding process.}
    \label{visual_att_graph}
    \end{figure}

    The visualization analysis in
    Section 4.6 ``Generated Ending Examples and Attention Visualization" is based on the attention weights ($\alpha_{h_{k,j}}^{(i)}$ and $\alpha_{x_{k,j}}^{(i)}$), as presented in Figure \ref{visual_att_graph}. Similarly we take as example the graph attention method to represent commonsense knowledge.

    The figure illustrates how the incremental encoding scheme with the multi-source attention  utilizes context clues and implicit knowledge.
    \\
    1) The left column:
    for \textbf{utilizing context clues}, when the model encodes $X_2$, {\it cooking} in $X_1$ obtains the largest state attention weight ($\alpha_{h_{k,j}}^{(i)}$), which illustrates {\it cooking} is an important word (or event) for the context clue. Similarly, the key events in each sentence have largest attention weights to some entities or events in the preceding sentence, which forms the context clue (e.g., {\it perfects} in $X_3$ to {\it right} in $X_2$, {\it lay/down} in $X_4$ to {\it perfect/everything} in $X_3$, {\it get/back} in $Y$ to {\it lay/down} in $X_4$, etc.).
    \\
    2) The right column:
 for the use of \textbf{commonsense knowledge}, each sentence has attention weights ($\alpha_{x_{k,j}}^{(i)}$) to the knowledge graphs of the preceding sentence (e.g. {\it eat} in $X_2$ to {\it meal} in $X_1$, {\it dinner} in $X_3$ to {\it eat} in $X_2$, etc.). In this manner, the knowledge information is added into the encoding process of each sentence, which helps story comprehension for better ending generation (e.g., {\it kitchen} in $Y$ to {\it oven} in $X_2$, etc.).

\end{document}